%% file: root.tex
\documentclass[letterpaper, 10 pt, conference]{ieeeconf}  % Comment this line out if you need a4paper

\IEEEoverridecommandlockouts                              % This command is only needed if 
\usepackage{graphics} % for pdf, bitmapped graphics files
\usepackage{epsfig} % for postscript graphics files
\usepackage{mathptmx} % assumes new font selection scheme installed
\usepackage{times} % assumes new font selection scheme installed
\usepackage{amsmath} % assumes amsmath package installed
\usepackage{amssymb}  % assumes amsmath package installed
\usepackage{booktabs} 
\usepackage{multirow}

\usepackage{subcaption}
\usepackage{makecell}   % \makecell (line breaks in cells)

\usepackage{graphicx}  % \resizebox, \includegraphics

\usepackage{capt-of}   % \captionof{table}{...}

\usepackage[dvipsnames]{xcolor}

\usepackage{algorithm}
\usepackage{algpseudocode}

\usepackage{makecell}
\usepackage{pifont}
\usepackage{siunitx}
\usepackage{hyperref}
\usepackage{cleveref}

\title{\LARGE \bf
Learning Locomotion on Discrete Terrain \\via Minimal Proximity Sensing
}

\author{
Jiale Fan, Connor Flynn, Tianao Xu, Junzhe He,
Andrei Cramariuc, Marco Hutter, and Robert Baines%
\thanks{All authors are with the Robotic Systems Lab, ETH Zurich.
E-mail: {\tt\small jialfan@ethz.ch}}%
\thanks{This work was supported by an ETH RobotX research grant funded through the ETH Zurich Foundation, the Branco Weiss Fellowship - Society in Science, administered by ETH Zurich, and the European Union’s Horizon Europe Framework Programme under grant agreement No 101121321. This work was conducted as part of ANYmal Research, a community to advance legged robotics. }%
}
\begin{document}

\maketitle
\thispagestyle{empty}
\pagestyle{empty}

%%%%%%%%%%%%%%%%%%%%%%%%%%%%%%%%%%%%%%%%%%%%%%%%%%%%%%%%%%%%%%%%%%%%%%%%%%%%%%%%

\begin{abstract}
Learning-based control has revolutionized dynamic locomotion, yet navigating unstructured terrain remains limited by a robot's incomplete awareness of imminent ground contact. While global perception systems such as LiDARs and depth cameras provide environmental context, they are frequently plagued by latencies, occlusions, and the high computational cost of dense geometric reconstruction. On the other hand, proprioceptive feedback is purely reactive, initiating corrections only after impact has occurred. 
This work explores embedding a minimal suite of low-cost, high-frequency infrared proximity sensors directly into the feet of a quadrupedal robot. These sensors provide ``pre-contact'' feedback that is robust to self-occlusions and significantly less computationally demanding than conventional vision-based pipelines.
By integrating these localized signals into a reinforcement learning framework, we enable the robot to anticipate terrain discontinuities such as gaps and stepping stones that are problematic for traditional perception stacks due to occlusions or state estimation drift. 
We demonstrate that such sparse, near-field sensing can be reliably modeled in simulation and transferred to the real world with high fidelity. 
Experimental results show that local proximity sensing substantially improves traversal robustness over discrete terrain and offers a low-power, low-latency alternative or complement to complex global perception suites in unpredictable environments.
For more information about results and methods, please see the project website: \url{https://sites.google.com/view/foot-tof/home}. 
\end{abstract}

\input{chapters/introduction}

\input{chapters/implementation}

\input{chapters/evaluations}

\input{chapters/hardware_deployment}

\input{chapters/conclusion}


%%%%%%%%%%%%%%%%%%%%%%%%%%%%%%%%%%%%%%%%%%%%%%%%%%%%%%%%%%%%%%%%%%%%%%%%%%%%%%%%

%%%%%%%%%%%%%%%%%%%%%%%%%%%%%%%%%%%%%%%%%%%%%%%%%%%%%%%%%%%%%%%%%%%%%%%%%%%%%%%%

%%%%%%%%%%%%%%%%%%%%%%%%%%%%%%%%%%%%%%%%%%%%%%%%%%%%%%%%%%%%%%%%%%%%%%%%%%%%%%%%


% \section*{ACKNOWLEDGMENT}

% The preferred spelling of the word ÒacknowledgmentÓ in America is without an ÒeÓ after the ÒgÓ. Avoid the stilted expression, ÒOne of us (R. B. G.) thanks . . .Ó  Instead, try ÒR. B. G. thanksÓ. Put sponsor acknowledgments in the unnumbered footnote on the first page.

%%%%%%%%%%%%%%%%%%%%%%%%%%%%%%%%%%%%%%%%%%%%%%%%%%%%%%%%%%%%%%%%%%%%%%%%%%%%%%%%

\bibliographystyle{IEEEtran}
\bibliography{IEEEabrv, main}

\clearpage
\onecolumn
% \section*{Supplementary Information \\ for \textit{Learning Locomotion on Discrete Terrain via Minimal Proximity Sensing}}
% \input{chapters/appendix}

\end{document}

%% file: chapters/introduction.tex
\section{Introduction}

% Start broad: robot performance and learning-based control improves performance 
Advances in learning-based control have enabled legged robots to engage in increasingly dynamic behaviors, including agile locomotion, object interaction, and whole-body manipulation~\cite{cheng2024extreme,hoeller_anymal_2024,lin2024locoman, ma2025learning}. 
%Indicate the challenge: perception of subtle, or oncoming terrain features underfoot 
Yet, the performance of most legged systems remains limited by incomplete awareness of impending contact with the terrain. 
Decisions about foothold placement, impact timing, and impact force modulation are often made using inferred or delayed information about contact onset. 
This lack of environmental awareness is particularly detrimental in unstructured environments with terrain discontinuities, such as gaps, edges, or protrusions. Locomotion failure may occur if such features are not detected at an appropriate spatial and temporal scale~\cite{nguyen2022continuous, vogel2025robust, kim2025highspeed}. 

\begin{figure}
    \centering
    \includegraphics[width=\linewidth]{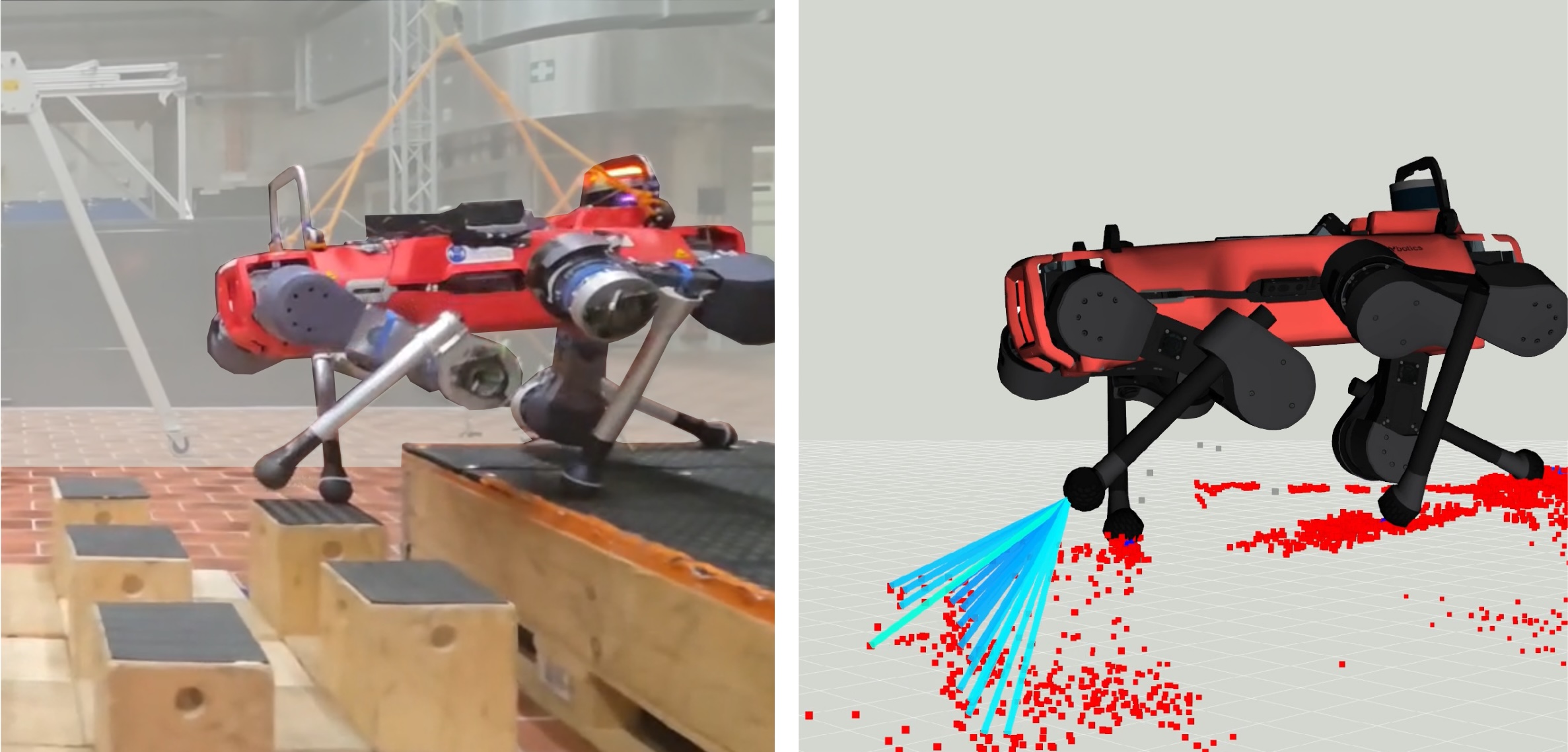}
    \caption{In this work, we integrate low-cost proximity sensors into the feet of a quadruped robot, granting it the ability to screen underfoot terrain with fewer occlusion issues compared to conventional perception-based state estimation. We study how the minimal sensing arrangement informs learned control policies for locomotion over discrete terrain, finding that proximity sensing facilitates reliable, noise-resistant locomotion policies.}
    \label{fig:pull_figure}
\end{figure}
% Overview of the various state estimation techniques common in quadrupeds 
Legged robots frequently rely on cameras, depth sensors, and proprioceptive joint measurements (e.g., torques, velocities, tracking accuracy) to infer information about themselves and the environment. 
Depth sensors (LiDAR or RGB-D cameras) provide rich global context but are susceptible to (self or environment) occlusions, limited fields of view, and time synchronization issues. 
State estimation can serve to create more complete and accurate 3D reconstructions of the environment. However, most state estimation methods are susceptible to small errors and latency issues~\cite{hoeller2022neural, miki2024learning, hoeller_anymal_2024, rudin2025parkour}. 
Limitations of existing state estimators are particularly pronounced during highly dynamic maneuvers, where drift and time-synchronization issues become more pronounced~\cite{bloesch2017iterated}. Additionally, with sensors mounted on the robot's chassis, limbs can self-occlude incoming or current contacts, making direct estimation from sensor data challenging~\cite{zhong2023touching, patel2020deep}.

%Sensor fusion + learning to infer contact events
Complementary approaches leverage filter-based state estimation using inertial measurements and contact constraints to explicitly model the robot’s motion and contact state~\cite{haddadin2008collision,bloesch2013state,camurri2017prob,grandia2023perceptive}. Modern reinforcement learning (RL)-based control policies take raw sensor observations as input to implicitly infer latent state and contact information~\cite{miki_learning_2022}. 
Teacher policies in simulation, granted full access to ground-truth terrain geometry and contact states, can acquire highly dynamic skills with impressive agility. 
However, when distilled to student policies constrained to onboard sensing and subject to real-world noise, performance degrades because existing onboard sensing modalities provide no direct, spatially resolved measurement of imminent contact~\cite{hoeller_anymal_2024, rudin2025parkour}.

% Acknowledge contact sensors
Contact sensors, such as pressure-sensitive resistors, have also been integrated into quadruped feet to provide explicit signals that facilitate better terrain state estimation~\cite{fondahl2012adaptive}. 
However, contact sensing at the moment of impact does not provide the necessary in-advance information about local terrain geometry. For dynamic foothold selection and gap traversal, anticipating contact before impact is essential. 

Moreover, many simulators struggle to realistically model high-frequency contact forces~\cite{makoviychuk2021isaac, mittal2025isaac}. Unlike contact sensing, which makes necessary simplifying assumptions, proximity sensing is a straightforward sensor to simulate during training because it uses hardware-accelerated ray casting.

% Proximity sensors: most related to our work 
Short-range proximity and time-of-flight (ToF) sensors may thus offer a solution to this state estimation quandary. Proximity sensors have been outfitted onto tabletop manipulators to permit body-scale collision detection and reactive behaviors~\cite{tsuji2019proximity, sander2025estimating,moon2021realtime}, as well as on small wheeled~\cite{pleterski2023miniature} or legged~\cite{sato2022robot} robots. 
Even so, the influence of proximity sensing in learning-based, whole-body legged locomotion over unstructured terrain remains relatively underexplored. 

% Explain how we fill the literature gap
Herein, we investigate lightweight, low-cost proximity sensors (that use infrared wavelengths) embedded directly into the feet of a quadruped robot (Fig.~\ref{fig:pull_figure}). We develop sensing units that are integrated into existing modular leg hardware with minimal modifications. By scanning a grid of points under the robot's foot prior to contact, the sensors provide local geometric awareness independent of global line-of-sight constraints. Proximity measurements are fed as observations into a learning-based control policy to generate emergent, agile, and safe terrain-seeking maneuvers. The contributions of this work are as follows:

%Enumerate contributions in typical Marco style: 
\begin{itemize}
\item The design and deployment of low-cost, readily integrable proximity sensing modules for the feet of quadruped robots that facilitate locomotion in discrete terrain environments. 

\item Characterization of the proximity sensor to obtain models of sensor noise, latency, and failure modes, which are in turn implemented in simulation training environments to bridge the sim-to-real gap; furthermore, evaluations of how a learned locomotion policy reacts to perception and model inaccuracies across different perceptive sensor sources.

\item Validation of the proposed sensing framework through real-world experiments, showcasing reliable traversal of discontinuous terrain.
\end{itemize}

% Summary statement to close the introduction about the significance of the work, and transition to the methods: 
Overall, this work interfaces a minimal sensing paradigm with learning-based control for legged locomotion. Rather than relying on complex global perception stacks, we find that strategically placed, task-relevant local sensors constitute a scalable component of systems intended for operation in real-world, unstructured environments.

%% file: chapters/implementation.tex
\section{Implementation}

%%%%%%%%%%%%%%%%%%%%%%%%%%%%%%%%%%%%%%%%%%%%%%%%%%%%%%%%%%

\subsection{Sensorized robotic feet}
\label{sec:sensors}
\begin{figure}[!t]
    \centering
    \includegraphics[width=.9\linewidth]{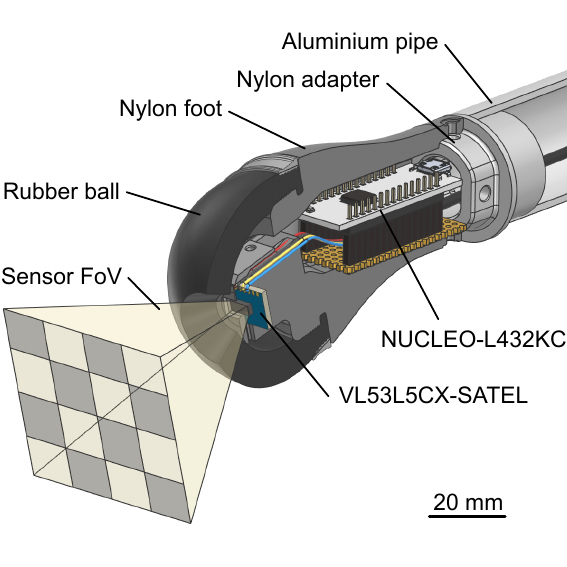} % there is also a version with underlined labels (add "_underlined")
    \caption{The proximity sensor and electronics are integrated directly into the foot. The grid shows the sensor Field-of-View and resolution.}
    \label{fig:hardware_figure}
\end{figure}

The feet of a commercially available ANYmal-D robot (ANYbotics) were modified to house an STMicroelectronics VL53L5CX ToF proximity sensor and the corresponding signal-processing electronics. The sensor has a \SI{45}{\degree} $\times$ \SI{45}{\degree} Field of View (FoV), split into a $4 \times 4$ grid, which corresponds to the sensor's resolution (see Fig. \ref{fig:hardware_figure}). We configure the sensor to use a resolution of $4 \times 4$ to obtain data at \SI{60}{\hertz},
to match the motion controller's frequency described in Sec.~\ref{sec:training}. Data is relayed from the sensor to the robot via a USB buffer with a measured delay of 20-\SI{60}{\milli\second}.

\begin{figure}[!t]
    \centering
    \includegraphics[width=1.0\linewidth]{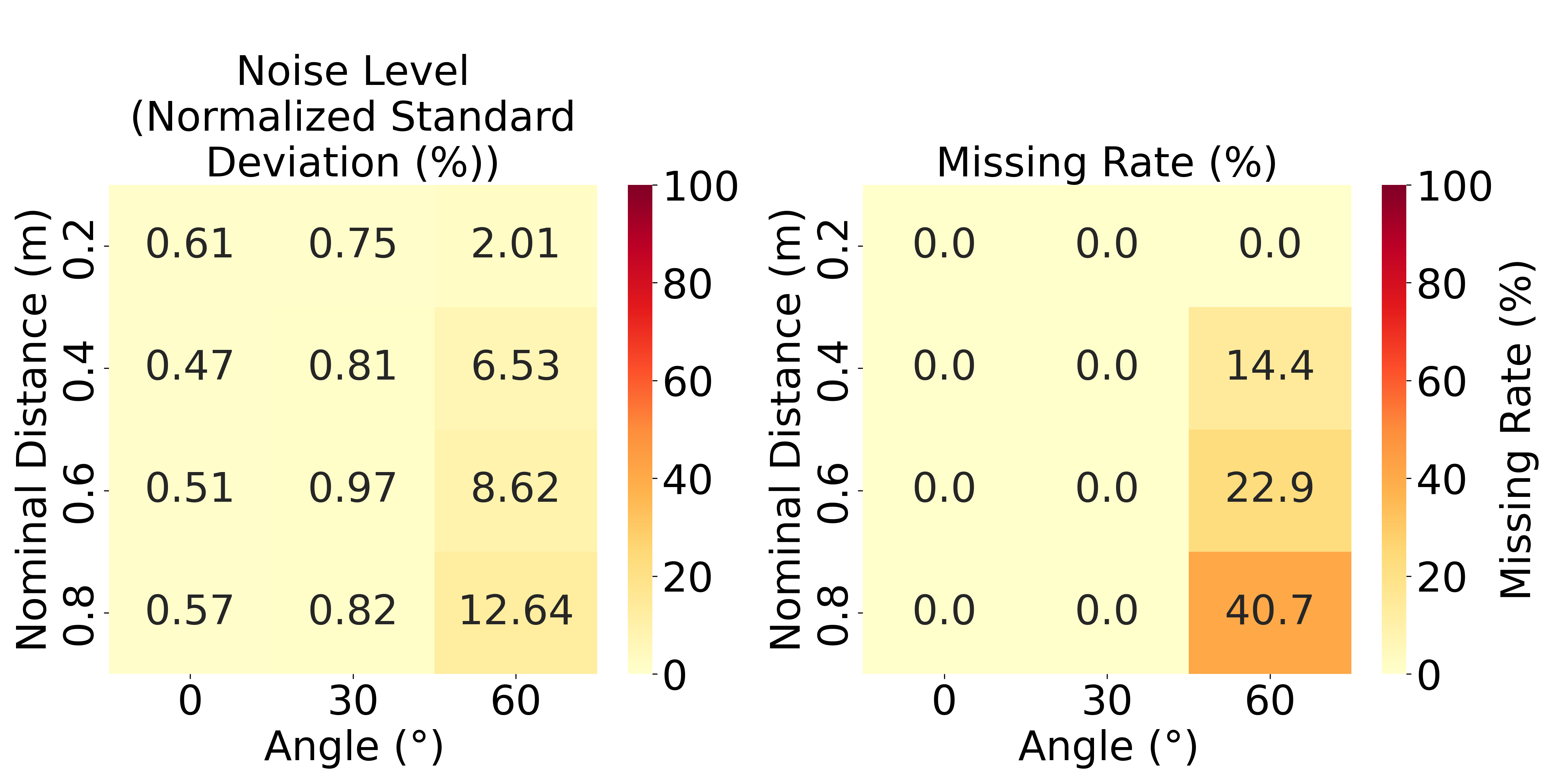} 
    \caption{Heatmaps of noise level and missing rate of the Time-of-Flight proximity sensor. The nominal distance denotes the distance from the sensor to the ground along an imaginary ray that is central in the sensor’s field of view, at the given incidence angle.}
    \label{fig:tof_heatmap}
\end{figure}

\begin{figure*}
    \centering
    \includegraphics[width=0.99\linewidth]{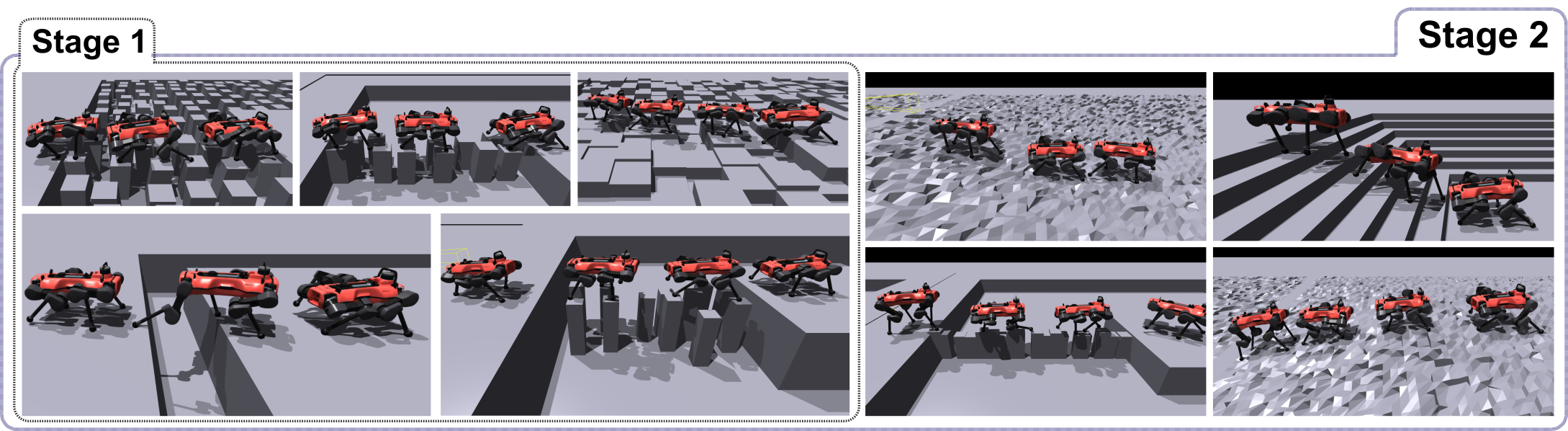}
    \caption{Training terrain examples. Stage 1 training comprises five terrain types: dense grid stones, two-row stones, rough box fields, platform–gap crossings, and mixed-row stones. Stage 2 includes all terrain types present in Stage 1 at elevated difficulty levels and further introduces four additional terrain types: balance beams, stairs, and both upward and downward pyramid slopes. }
    \label{fig:terrains}
\end{figure*}

Unlike other commonly-used depth sensors that output discrete point measurements, the VL53L5CX evenly covers the entire FoV with only 16 cells. For each cell in the $4 \times 4$ grid, the sensor reports the minimum distance to an obstacle within that cell. Each zone of the sensor grid provides a range value with a resolution of \SI{1}{\milli\meter} and a binary measurement-validity flag. Sensor readings above \SI{1}{\meter} are clipped because beyond this value, noise increases and the relevance to planning locomotion footholds decreases.

% Refer to Tab. \ref{tab:sensor_configuration} for the sensor configuration.
% Under ideal conditions the specified maximum range is \SI{4}{\meter} with \SI{\pm5}{\percent} accuracy. In practice, performance depends strongly on surface reflectivity, texture, and ambient lighting. In our experiments, reliable measurements were obtained up to approximately \SI{1}{\meter}. Beyond this, the number of valid returns decreased significantly. 

% #RB: commenting out because sufficient info in above description. But we can keep as commented out in here. 
% \begin{table}[h]
%     \centering
%     \caption{Sensor Configuration}
%     \label{tab:sensor_configuration}
%     \begin{tabular}{ll}
%     \toprule
%     Parameter & Value \\
%     \midrule
%     Resolution & 16 ($4 \times 4$) \\
%     Frequency & \SI{60}{\hertz} \\
%     Ranging Mode & Autonomous \\
%     Integration Time & \SI{1}{\milli\second} \\
%     Target Order & Closest \\
%     Sharpener & \SI{0}{\percent} \\
%     \bottomrule
%     \end{tabular}
% \end{table}

% \subsection{Integration into robot foot}

The proximity sensor was placed snugly inside the rubber ball foot with a machined aperture for the FoV (Fig. \ref{fig:hardware_figure}). The optical axis is tilted approximately \SI{15}{\degree} outward relative to the sagittal plane to expand coverage of the anticipated foothold region during a standard foot swing. 

Application of the ToF sensor on the bottom of a quadruped's feet requires it to operate over a wide range of distances and incidence angles while stepping on different surfaces. To characterize sensor reliability under these conditions, we collected measurements at different distances and angles on representative deployment surfaces. 
We found that the sensor is highly accurate and reliable at short range (e.g., under \SI{0.6}{\meter}) and when oriented nearly perpendicularly to the target surface (Fig.~\ref{fig:tof_heatmap}). 
However, greater distances and oblique incidence angles cause increased measurement noise and missing data. We use the data collected from this experiment as the basis for the sensor-noise model in simulation.

% Removed content: 

%ToF sensing measurements can be inaccurate or  invalid (e.g., no reading returned). 
% While a breakout-based implementation was chosen for rapid prototyping, the sensor and microcontroller could be placed together onto a custom PCB to reduce the footprint and enable more versatile multi-sensor configurations.
% The microcontroller operates as a USB CDC device and streams measurements directly to the main controller on power-up, letting the operating system handle the communication. The feet operate asynchronously and the controller is immediately served the newest measurement.

% End-to-end latency is dominated by the \SI{60}{\hertz} sampling period, USB communication, and aliasing with the \SI{50}{\hertz} control loop, resulting in an average delay of \SI{80+-8}{\milli\second}. This can be significantly reduced by using a more sophisticated communication protocol and synchronizing the sampling with the control loop.

%%%%%%%%%%%%%%%%%%%%%%%%%%%%%%%%%%%%%%%%%%%%%%%%%%%%%%%%%%

\subsection{Control methods}
\label{sec:training}

\subsubsection{Overview}

Leveraging foot-centric proximity sensor feedback, we propose a minimal approach for learning robust locomotion policies over discrete terrain. The approach is characterized by:
\begin{enumerate}
    \item No reliance on a standalone map-generation stack (e.g., \cite{miki2022elevation, dong2025marg}), which typically requires extensive tuning and is highly sensitive to noise.
    \item No need for dedicated map-processing mechanisms to handle the high dimensionality of elevation maps, which often involve computationally expensive operations such as multi-head attention (e.g., \cite{he2025attention, yu2026start}).
    \item A single unified policy for all terrain types without relying on terrain-specialist training (e.g., \cite{zhang2024learning}).
\end{enumerate}
The policy directly takes as observations the low-dimensional foot proximity sensor readings, implemented in simulation as ray casters mounted on the robot's feet with an analogous FoV to those of the real sensors.
We simulate the large FoV of the beams by ray casting at a higher resolution (ray casted towards the 4 corners and the center of each sensor zone) and then using linear interpolation to approximate the closest point in each of the $4\times 4$ grid cells. 
In addition to ToF readings, other proprioceptive inputs including joint position, gravity, body linear and rotational velocities are given to the network.
The policy is implemented as a Long Short-Term Memory (LSTM) network~\cite{hochreiter1997long} trained with Proximal Policy Optimization (PPO)~\cite{schulman2017proximal, schwarke2025rsl} entirely in the Isaac Gym simulation environment~\cite{rudin2022learning}. The robustness of the proposed method is examined in Sec.~\ref{sec:robustness}.

\subsubsection{Training stages, curriculum and terrains}
Training is conducted in two stages. Stage 1 primarily lets the robot familiarize itself with the terrain types and learn the contexts associated with different sensory inputs.
Stage 1 training employs a terrain composition comprised mostly of stepping stones. For each terrain type, a curriculum progressively increases in difficulty within each terrain type by interpolating geometric parameters from easy to hard settings (Fig.~\ref{fig:terrains}). 
As the curriculum progresses, the foothold size decreases, inter-foothold spacing increases, and vertical irregularity increases.
By the end of Stage 1, the robot should learn to interpret the perceptual signals, develop basic locomotion reflexes, and traverse all terrains under ideal sensing conditions. 

Stage 2 includes training terrains from Stage 1, but these terrains are shifted into a more challenging regime from the outset (e.g., stones are constrained to smaller size bounds with larger minimum gaps and height variations), while the variety of initial robot poses is also expanded. Additional terrain types, like stairs and faceted ground (see Fig.~\ref{fig:terrains}), are introduced to promote generalization to environments that require diverse gaits and walking styles. 
% while also suppressing undesirable behaviors such as foot dragging. 
By the end of the two training stages, a single unified policy is obtained that can traverse all target terrain types.

\begin{figure*}[h!]
    \centering
\includegraphics[width=0.98\textwidth]{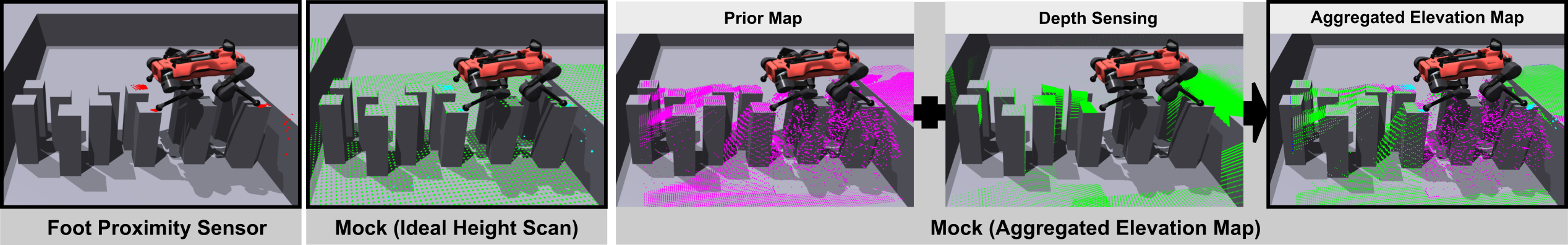} 
    \caption{Examples of observation configurations used in Sec.~\ref{sec:robustness}.
    \textbf{A.} \textit{Foot Proximity Sensor.} Foot proximity sensor models are instantiated as ray casters, with ray hit points shown in \textcolor{red}{red}.
    \textbf{B.} \textit{Mock (Ideal Height Scan)}. Access to an ideal, noise-free egocentric elevation map (\textcolor{green}{green}) is assumed, from which foot proximity sensor readings (\textcolor{cyan}{cyan}) are emulated.
    \textbf{C.} \textit{Mock (Aggre. Elevation Map)}. We further emulate an elevation-mapping pipeline constructed from depth images (\textcolor{green}{green}) acquired by six chassis-mounted depth cameras. The prior elevation map (\textcolor{magenta}{magenta}) from the previous update is transformed into the current base frame using estimated odometry to compensate for the cameras’ blind zones. Foot proximity sensor readings (\textcolor{cyan}{cyan}) are then emulated based on the aggregated elevation map. }
    \label{fig:variants}
\end{figure*}

% Gap terrains likewise progress from narrow to wide separations (approximately 0.05–0.8m).

% \begin{table}
%   \centering
%   \caption{Training Settings}
%   \label{tab:training settings}
%   \begin{tabular}{l|l}
%     \hline
%     \textbf{Name} & \textbf{Equation} \\ \hline

%   \end{tabular}
% \end{table}

\subsubsection{Reward function}
The robot continuously observes distance and direction to a specified target position in each episode.
The reward function comprises sparse task rewards provided only at the end of each episode, dense task rewards that offer strong guidance during learning, and regularization rewards (following \cite{zhang2024learning} with modifications) that encourage efficient use of limited energy and suppress undesirable behaviors. Regularization rewards mitigate abnormal postures after reaching the target position, excessive jumping, stagnation, or overly aggressive gaits. For more detailed information about reward formulation, see the supplementary document on the project website.

\subsubsection{Sensor-specific randomization}

We strive to model the degradation characteristics of the proximity sensors to make policies trained in simulation more readily transferable to the real-world quadruped platform. 
Namely, measurement inaccuracies are modeled as a static offset, proportional noise, and Gaussian noise during training.
Missing distance readings are modeled by randomly marking a fraction of readings as missing at each timestep and replacing them with zeros. 
Finally, we explicitly introduce temporal delays in the foot proximity sensor readings by delaying the measurements for a few steps, which corresponds to the real-world delays measured in Sec.~\ref{sec:sensors}.

For each degradation type, at the beginning of each episode, the randomization level is sampled uniformly from $[0,\text{upper bound}]$. This procedure ensures that the full spectrum of sensing conditions--ranging from ideal perception to severely corrupted observations--is encountered, reducing the risk of overfitting to any specific degradation level. The degradation level is kept very low in Stage 1 to facilitate rapid skill acquisition, and is then increased in Stage 2 to mirror real-world levels.

%% file: chapters/evaluations.tex
\begin{figure*}
    \centering
\includegraphics[width=0.98\textwidth]{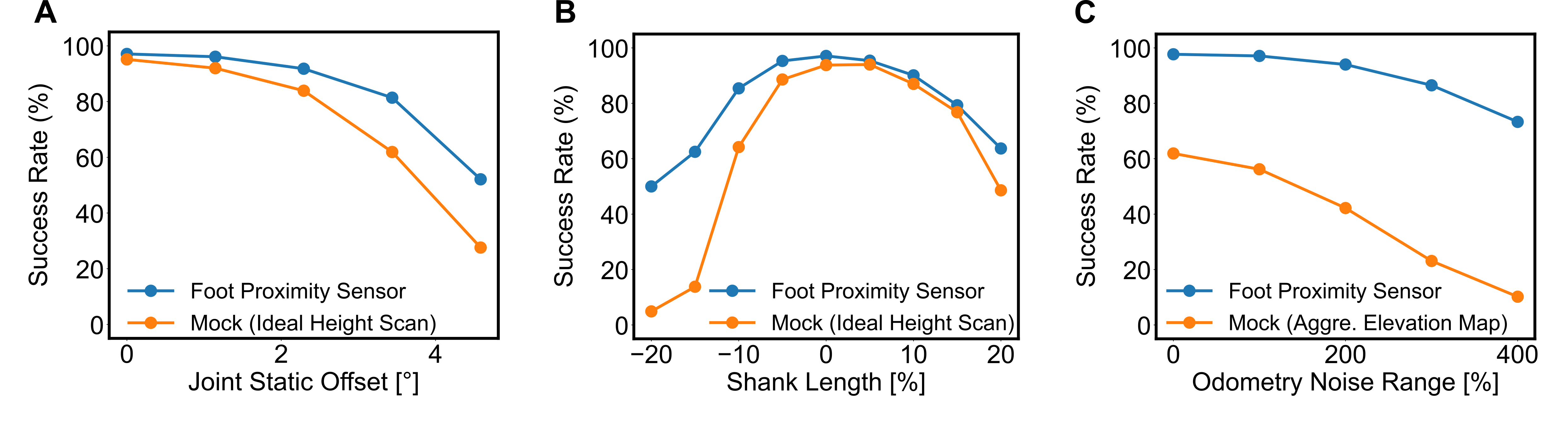} 
    \caption{Robustness assessment: success rate on an ensemble of discrete terrain traversal tasks as a function of different levels and kinds of noise injected into the robot simulation (out of training distribution of the policy). Across all the conditions \textbf{A.} static joint position offset, \textbf{B.} robot shank length, and \textbf{C.} odometry noise, the foot proximity sensor performs better than the robocentric baselines.}
    \label{fig:robustness}
\end{figure*}

\section{Robustness Evaluation}
\label{sec:robustness} 

As benchmarks for comparison, we investigated whether imminent contact information could be effectively recovered from conventional, base-centric sensors as an alternative to the foot-centric proximity pipeline. 
We conducted experiments in simulation to remove biases arising from varying real-world depth noise and the sensor-specific tuning often required during training.

Specifically, we evaluated locomotion performance across two configurations: \textit{Mock (Ideal Height Scan)} and \textit{Mock (Aggre. Elevation Map)}. In these variants, the ``pre-contact'' signals are not measured directly at the feet but are instead kinematically projected from global robocentric observations (Fig. \ref{fig:variants}). The \textit{Mock (Ideal Height Scan)} is a top-down ray caster in simulation that projects only against the terrain. This provides a $(5 \mathrm{m} \times 5 \mathrm{m})$ 2.5D scan with a spatial resolution of 5cm, and no depth noise. In \textit{Mock (Aggre. Elevation Map)}, we further emulate the pipeline of iteratively constructing the height scan from depth images acquired by six chassis-mounted depth cameras. At each timestep, cells within the FoV of all depth cameras are updated directly from depth measurements, whereas cells in the dead zones instead rely on the prior map transformed to the current frame. We assume that this transformation is obtained via zero-order integration of the estimated base linear and angular velocities.

With these two benchmarks, we effectively isolate the benefits of local, direct sensing from the inaccuracies inherent in estimating near-foot terrain geometry from a distal, body-fixed perspective. 
Height scans are sampled from \textit{Mock (Ideal Height Scan)} and \textit{Mock (Aggre. Elevation Map)} near the feet to generate data analogous to foot-mounted proximity sensors. 
The consistent data structures permit the use of the same locomotion policy across all evaluations, ensuring that performance comparisons reflect only the differences in sensor input modality. 
Since all ray casters in simulation cast only against the terrain, not the robot, there are no self-occlusions. This enforces a fair comparison between the proximity and benchmark sensor modalities, as we do not have to implement arbitrary rules about how to handle occlusions.

Performance is evaluated by the success rate over 1000 runs on a challenging test terrain combination composed of 80\% stepping-stone terrain and 20\% gap terrain, with difficulty levels corresponding to 90\% curriculum progress for their respective terrain types in Stage 1. A run is considered successful only if the robot reaches a commanded position within \SI{0.5}{\meter} by the end of the episode; otherwise, it is classified as a failure. The same locomotion policy, trained without any of the tested noise models presented below, is used for all evaluation runs. All ray caster distances return are noise-free. The robustness evaluation is conducted in the Isaac Gym simulator.

\subsection{Robustness to static joint position offset}
\label{sec:joint_offset}

Static joint position reading offsets occur when a fixed value is added to the position readings of all joints. This significantly differs from the standard Gaussian noise added to the joint readings in training. We compared the performance of \textit{Foot Proximity Sensor} and \textit{Mock (Ideal Height Scan)} policies under this perturbation in Fig.~\ref{fig:robustness}A. When no static offset is present, \textit{Mock (Ideal Height Scan)} performs on par with foot-centric sensing. However, once a static bias is introduced into the joint readings, the performance of base-centric sensing diminishes significantly faster than that of foot-centric sensing. Performance reduces because the biased joint measurements corrupt the estimated foot positions used to query the proximity readings.

\subsection{Robustness to robot model dimension mismatches}

Model mismatch (differences between a robot's real body and the simulated body on which the policy was trained) can stem from inaccuracies in hardware mounting, manufacturing tolerances, or wear. We investigated a representative case of model mismatch by varying the robot's shank length by up to $\pm 20 \%$ while keeping its mass and inertia constant (Fig.~\ref{fig:robustness}B). 
Both the policy and the kinematics-based proximity retrieval module continue to assume the nominal robot model.
Under this perturbation, \textit{Foot Proximity Sensor} demonstrated greater robustness than \textit{Mock (Ideal Height Scan)} in the presence of shank length mismatch. Similar to the case of static joint offsets, the performance of \textit{Mock (Ideal Height Scan)} declines due to shifts in the estimated foot positions caused by the unmodeled shank lengths, which in turn cause errors in estimating proximity to the ground.

\subsection{Robustness to odometry noise}

Acquiring an elevation map on a quadruped typically relies on a sophisticated perception stack that processes depth camera streams (e.g.,~\cite{miki2022elevation}). Due to blind zones and sensor occlusions, elevation maps from previous time steps must be aggregated into the current base frame using odometry. However, odometry for legged systems is prone to drift, and Simultaneous Localization and Mapping systems are challenging to deploy due to the controller's real-time requirements.

We treat the commonly-used velocity estimation noise in training (linear velocity: uniformly within $\pm 0.1\,\mathrm{m/s}$ along each axis in the base frame; angular velocity: uniformly within $\pm 0.2\,\mathrm{rad/s}$) as the $100\%$ odometry noise level, and evaluate performance under noise levels ranging from $0\%$ to $400\%$ (Fig.~\ref{fig:robustness}C). 
Results show that the degradation in \textit{Mock (Aggre. Elevation Map)} remains substantially more pronounced than that of \textit{Foot Proximity Sensor}, since the success rate drop from noise level 0\% to 400\% is 51.7\% to \ 24.3\%).

% Even in the absence of odometry noise, \textit{Mock (Aggre. Elevation Map)} underperforms relative to \textit{Mock (Ideal Height Scan)}, highlighting the disparity between a naive elevation mapping baseline and a well-tuned pipeline. 
% Even without self-occlusions from the feet blocking the view of contact points, the robocentric sensor setup struggles to obtain an equally clean representation of the environment. 

\subsection{Summary}

Across all three types of evaluations, in the absence of injected degradations, base-centric sensing models achieve reasonable performance. However, once perturbations were introduced, the performance of base-centric sensing deteriorated markedly faster than that of foot-centric sensing. These results suggest that foot-centric sensing constitutes a viable strategy for narrowing the gap between simulation and real-world behaviors. 
More broadly, the results underscore that sensing as close as possible to the point(s) of interest may be the most effective sensing strategy to obtain clear and reliable environmental information. 
Adding intermediary components based on other proprioceptive or exteroceptive components increases the number of failure points and the sensitivity to noise.

\begin{figure*}[htb] 
    \centering
    \includegraphics[width=0.99\linewidth]{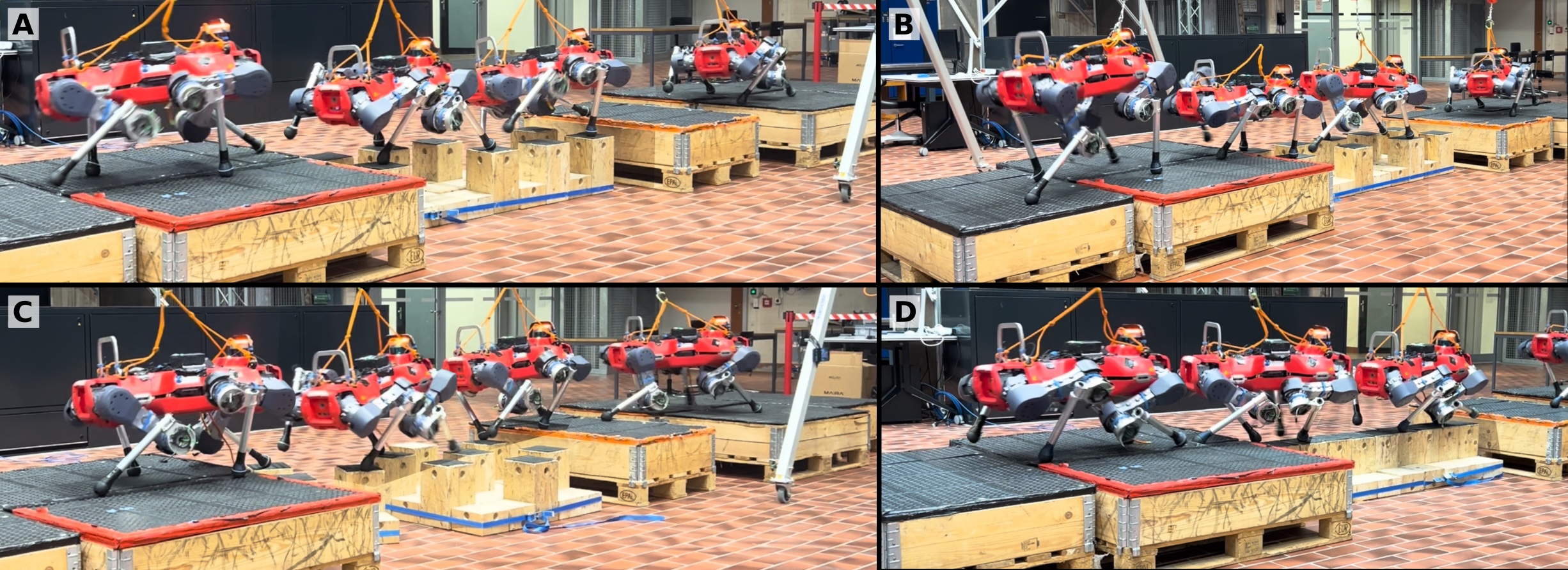} 
    \caption{Deployment examples using only the foot proximity sensors on different stepping stone configurations: \textbf{A.} Sparse double-row; \textbf{B.} Regular double-row; \textbf{C.} Scattered; \textbf{D.} Balance beam.}
    \label{fig:deployment_composites}
\end{figure*}

%% file: chapters/hardware_deployment.tex
\section{Real-World Deployment}

\begin{figure*}[t] 
    \centering
    \includegraphics[width=1.0\textwidth]{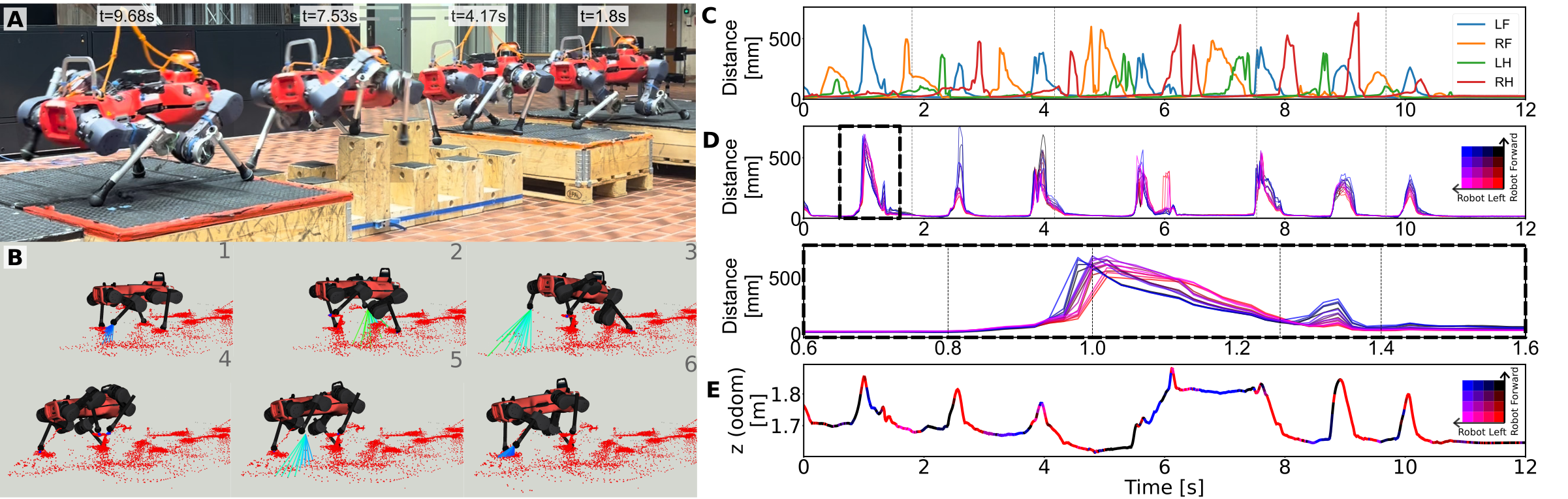} 
    \caption{Hardware deployment. \textbf{A.} Composite image of the robot traversing gaps and stepping stones of varying heights.
\textbf{B.} Snapshots from rosbag playback during traversal of an elevated stone, with current foot proximity sensor rays and historical ray hit points visualized.
\textbf{C.} Proximity sensor data for each foot, averaged over the 16 individual channels per foot. \textbf{D.} Raw data from the LF foot sensor. The 16 individual zones from the ToF sensor are plotted with different colors to emphasize their relative order at each timestep. A zoom-in view of the gap-traversal segment (0.6s-1.6s) is shown below the plot over whole episode. \textbf{E.} Z-coordinate of the LF foot with respect to the odometry frame plotted over time. The color denotes the index of the zone with the minimum reading at each timestep. }
    \label{fig:deployment_analysis}
\end{figure*}

% \begin{figure*}[htb] 
%     \centering
%     \includegraphics[width=0.94\linewidth]{figures/odom_z_colored.png} 
%     \caption{?? }
%     \label{fig:ddom_z_colored}
% \end{figure*}

% \begin{figure}[htb] 
%     \centering
%     \includegraphics[width=0.94\linewidth]{figures/closer_single_foot_tof.png} 
%     \caption{?? }
%     \label{fig:closer_tof}
% \end{figure}

\subsection{Evaluation setup}

We deployed our policy on an ANYmal-D robot (ANYbotics) on the following canonical sparse terrains:

\begin{enumerate}
    \item Horizontal gaps of up to \SI{60}{\centi\meter}.
    \item Aligned or scattered rows of stepping stones; each stone has a 20$\times$\SI{20}{\centi\meter} top area. A layer of supporting square bricks is used for stabilization. The bricks' top surfaces are \SI{20}{\centi\meter} lower than the regular stepping stones and can also serve as valid footholds. We also sporadically include \SI{20}{\centi\meter} higher stepping stones to vary the vertical profile. 
    \item A straight row of stones at the same height, forming a beam \SI{20}{\centi\meter} wide.
\end{enumerate}

%%%%%%%%%%%%%%%%%%%%%%%%%%%%%

\subsection{Performance}

The robot can traverse all terrain types with an average speed of 0.52~m/s and a balance of stability and agility (Fig.~\ref{fig:deployment_composites}). One representative run is shown in Fig.~\ref{fig:deployment_analysis}A, in which the robot sequentially traverses a gap, then a sparse stepping-stone section featuring elevated stones, and finally successfully reaches the target platform. Fig.~\ref{fig:deployment_analysis}B illustrates the process by which the robot detects an elevated stone, identifies its top surface, determines a suitable foothold, and sequentially places the left front and left hind feet onto the stone.

Fig.~\ref{fig:deployment_analysis}C presents the proximity readings averaged across all zones per foot over the entire run, providing an approximate representation of the terrain characteristics perceived by the robot at each moment. 
Further details are provided in Fig.~\ref{fig:deployment_analysis}D, which shows the readings from all zones of the LF foot proximity sensor. When we examine the gap-traversal segment (0.6–\SI{1.6}{\second}), several distinct phases are apparent, providing insight into the policy's reflexive behavior when equipped with foot-centric proximity sensing. First, between \SI{0.80}{\second} and \SI{1.00}{\second}, the foot is lifted, and the measured distances increase sharply, indicating that the robot becomes aware of the trench. As the foot continues to move (1.00–\SI{1.26}{\second}), a reversal in the relative ordering of the zone readings signals detection of the side surface of the next platform; the robot subsequently reaches toward it as the measured distances decrease. Finally, another reversal in the zone ordering (1.26–\SI{1.40}{\second}), together with a sudden increase in distance, indicates that the foot has approached the upper surface of the next platform, which naturally leads to a touchdown.

To provide further insight into the active perception behaviors learned by the policy from direct foot proximity inputs, Fig.~\ref{fig:deployment_analysis}E presents the z-coordinate of the LF foot with respect to the odometry frame over time, with color indicating the index of the zone exhibiting the minimum range measurement at each timestep. The identity of the minimum-reading zone can be interpreted as a coarse proxy for the surface normal direction of the prospective foothold, under the simplifying assumption that all ray intersections lie on a common planar surface. The results indicate that touchdown events are typically initiated only when a nearby, approximately horizontal surface is detected. Conversely, sustained detection of a quasi-vertical surface at a distance correlates with continued foot-lifting behavior, which ultimately guides the robot towards the elevated stone.

%% file: chapters/conclusion.tex
\section{Conclusion}

This work demonstrates that high-fidelity locomotion in complex environments does not strictly require high-bandwidth, computationally expensive vision pipelines. 
By integrating low-latency, foot-mounted infrared proximity sensors, we provide quadrupedal platforms with a pre-contact reflex that significantly improves reliability in environments where traditional sensors fail. 
As a result, our platform successfully navigated challenging tasks such as stepping-stone traversal and gap-crossing, where vision-based methods often struggle due to self-occlusion or perceptual aliasing. 
Our approach additionally yields emergent active perception behaviors, such as using the feet sensors to scan the environment. 
% In simulation, we also show that it is significantly more robust to many of the noise modalities that more complex perceptive systems struggle with. 
Despite these performance gains, several physical and environmental constraints remain that merit further investigation. 

The accuracy of ToF measurements depends on the optical properties of the substrate, and highly absorptive or specular surfaces can lead to signal attenuation and unreliable distance estimates. Furthermore, hardware longevity in the field remains a concern. Sensor apertures are susceptible to clogging by mud or debris during high-impact locomotion, suggesting a need for IR-transparent housings or self-clearing mechanisms. Future research will also focus on morphological optimization to determine the optimal density and placement of these sensors across the robot's limbs and undercarriage to provide a more holistic near-field awareness.
Overall, this work suggests that the path to truly robust autonomous robots may not only lie in ever-more complex algorithms, but in better-integrated, task-informed, and localized sensing.

%% file: main.bib
@INPROCEEDINGS{cheng2024extreme,
  author={Cheng, Xuxin and Shi, Kexin and Agarwal, Ananye and Pathak, Deepak},
  booktitle={2024 IEEE International Conference on Robotics and Automation (ICRA)}, 
  title={Extreme Parkour with Legged Robots}, 
  year={2024},
  volume={},
  number={},
  pages={11443-11450},
  doi={10.1109/ICRA57147.2024.10610200}}

@article{hoeller_anymal_2024,
	title = {{ANYmal} parkour: {Learning} agile navigation for quadrupedal robots},
	volume = {9},
	issn = {2470-9476},
	shorttitle = {{ANYmal} parkour},
	doi = {10.1126/scirobotics.adi7566},
	language = {en},
	number = {88},
	urldate = {2024-03-14},
	journal = {Science Robotics},
	author = {Hoeller, David and Rudin, Nikita and Sako, Dhionis and Hutter, Marco},
	month = mar,
	year = {2024},
	pages = {eadi7566},
}

@article{miki_learning_2022,
	title = {Learning robust perceptive locomotion for quadrupedal robots in the wild},
	volume = {7},
	issn = {2470-9476},
	doi = {10.1126/scirobotics.abk2822},
		language = {en},
	number = {62},
	urldate = {2022-03-23},
	journal = {Science Robotics},
	author = {Miki, Takahiro and Lee, Joonho and Hwangbo, Jemin and Wellhausen, Lorenz and Koltun, Vladlen and Hutter, Marco},
	month = jan,
	year = {2022},
	pages = {eabk2822},
}

@INPROCEEDINGS{lin2024locoman,
  author={Lin, Changyi and Liu, Xingyu and Yang, Yuxiang and Niu, Yaru and Yu, Wenhao and Zhang, Tingnan and Tan, Jie and Boots, Byron and Zhao, Ding},
  booktitle={2024 IEEE/RSJ International Conference on Intelligent Robots and Systems (IROS)}, 
  title={LocoMan: Advancing Versatile Quadrupedal Dexterity with Lightweight Loco-Manipulators}, 
  year={2024},
  volume={},
  number={},
  pages={6877-6884},
  doi={10.1109/IROS58592.2024.10801980}}

@ARTICLE{hoeller2022neural,
  author={Hoeller, David and Rudin, Nikita and Choy, Christopher and Anandkumar, Animashree and Hutter, Marco},
  journal={IEEE Robotics and Automation Letters}, 
  title={Neural Scene Representation for Locomotion on Structured Terrain}, 
  year={2022},
  volume={7},
  number={4},
  pages={8667-8674},
  doi={10.1109/LRA.2022.3184779}}

@INPROCEEDINGS{sander2025estimating,
  author={Sander, Jack and Caroleo, Giammarco and Albini, Alessandro and Maiolino, Perla},
  booktitle={2025 34th IEEE International Conference on Robot and Human Interactive Communication (RO-MAN)}, 
  title={Estimating Scene Flow in Robot Surroundings with Distributed Miniaturised Time-of-Flight Sensors}, 
  year={2025},
  volume={},
  number={},
  pages={493-499},
  doi={10.1109/RO-MAN63969.2025.11217813}}

@ARTICLE{tsuji2019proximity,
  author={Tsuji, Satoshi and Kohama, Teruhiko},
  journal={IEEE Sensors Journal}, 
  title={Proximity Skin Sensor Using Time-of-Flight Sensor for Human Collaborative Robot}, 
  year={2019},
  volume={19},
  number={14},
  pages={5859-5864},
  doi={10.1109/JSEN.2019.2905848}}

@ARTICLE{pleterski2023miniature,
  author={Pleterski, Jan and Škulj, Gašper and Esnault, Corentin and Puc, Jernej and Vrabič, Rok and Podržaj, Primož},
  journal={IEEE Transactions on Instrumentation and Measurement}, 
  title={Miniature Mobile Robot Detection Using an Ultralow-Resolution Time-of-Flight Sensor}, 
  year={2023},
  volume={72},
  number={},
  pages={1-9},
  doi={10.1109/TIM.2023.3318710}}

@INPROCEEDINGS{haddadin2008collision,
  author={Haddadin, Sami and Albu-Schaffer, Alin and De Luca, Alessandro and Hirzinger, Gerd},
  booktitle={2008 IEEE/RSJ International Conference on Intelligent Robots and Systems}, 
  title={Collision Detection and Reaction: A Contribution to Safe Physical Human-Robot Interaction}, 
  year={2008},
  volume={},
  number={},
  pages={3356-3363},
  doi={10.1109/IROS.2008.4650764}}

@ARTICLE{grandia2023perceptive,
  author={Grandia, Ruben and Jenelten, Fabian and Yang, Shaohui and Farshidian, Farbod and Hutter, Marco},
  journal={IEEE Transactions on Robotics}, 
  title={Perceptive Locomotion Through Nonlinear Model-Predictive Control}, 
  year={2023},
  volume={39},
  number={5},
  pages={3402-3421},
  doi={10.1109/TRO.2023.3275384}}

@ARTICLE{moon2021realtime,
  author={Moon, Seung Jae and Kim, Jinsol and Yim, Hongsik and Kim, Yeeun and Choi, Hyouk Ryeol},
  journal={IEEE Robotics and Automation Letters}, 
  title={Real-Time Obstacle Avoidance Using Dual-Type Proximity Sensor for Safe Human-Robot Interaction}, 
  year={2021},
  volume={6},
  number={4},
  pages={8021-8028},
  doi={10.1109/LRA.2021.3102318}}

@INPROCEEDINGS{vogel2025robust,
  author={Vogel, Dylan and Baines, Robert and Church, Joseph and Lotzer, Julian and Werner, Karl and Hutter, Marco},
  booktitle={2025 IEEE/RSJ International Conference on Intelligent Robots and Systems (IROS)}, 
  title={Robust Ladder Climbing with a Quadrupedal Robot}, 
  year={2025},
  volume={},
  number={},
  pages={7239-7244},
  doi={10.1109/IROS60139.2025.11247166}}

@INPROCEEDINGS{nguyen2022continuous,
  author={Nguyen, Chuong and Bao, Lingfan and Nguyen, Quan},
  booktitle={2022 IEEE 61st Conference on Decision and Control (CDC)}, 
  title={Continuous Jumping for Legged Robots on Stepping Stones via Trajectory Optimization and Model Predictive Control}, 
  year={2022},
  volume={},
  number={},
  pages={93-99},
  doi={10.1109/CDC51059.2022.9993259}}

@article{kim2025highspeed,
author = {Hyeongjun Kim  and Hyunsik Oh  and Jeongsoo Park  and Yunho Kim  and Donghoon Youm  and Moonkyu Jung  and Minho Lee  and Jemin Hwangbo },
title = {High-speed control and navigation for quadrupedal robots on complex and discrete terrain},
journal = {Science Robotics},
volume = {10},
number = {102},
pages = {eads6192},
year = {2025},
doi = {10.1126/scirobotics.ads6192},}

@ARTICLE{sato2022robot,
  author={Sato, Ryuki and Arita, Hikaru and Ming, Aiguo},
  journal={IEEE Access}, 
  title={Pre-Landing Control for a Legged Robot Based on Tiptoe Proximity Sensor Feedback}, 
  year={2022},
  volume={10},
  number={},
  pages={21619-21630},
  doi={10.1109/ACCESS.2022.3153127}}

@INPROCEEDINGS{fondahl2012adaptive,
  author={Fondahl, Kristin and Kuehn, Daniel and Beinersdorf, Frank and Bernhard, Felix and Grimminger, Felix and Schilling, Moritz and Stark, Tobias and Kirchner, Frank},
  booktitle={2012 4th IEEE RAS \& EMBS International Conference on Biomedical Robotics and Biomechatronics (BioRob)}, 
  title={An adaptive sensor foot for a bipedal and quadrupedal robot}, 
  year={2012},
  volume={},
  number={},
  pages={270-275},
  doi={10.1109/BioRob.2012.6290735}}

@ARTICLE{camurri2017prob,
  author={Camurri, Marco and Fallon, Maurice and Bazeille, Stéphane and Radulescu, Andreea and Barasuol, Victor and Caldwell, Darwin G. and Semini, Claudio},
  journal={IEEE Robotics and Automation Letters}, 
  title={Probabilistic Contact Estimation and Impact Detection for State Estimation of Quadruped Robots}, 
  year={2017},
  volume={2},
  number={2},
  pages={1023-1030},
  doi={10.1109/LRA.2017.2652491}}

@inproceedings{rudin2022learning,
  title={Learning to walk in minutes using massively parallel deep reinforcement learning},
  author={Rudin, Nikita and Hoeller, David and Reist, Philipp and Hutter, Marco},
  booktitle={Conference on robot learning},
  pages={91--100},
  year={2022},
  organization={PMLR}
}

@ARTICLE{yu2026start,
  author={Yu, Ruiqi and Wang, Qianshi and Li, Hongyi and Jun, Zheng and Wang, Zhicheng and Wu, Jun and Zhu, Qiuguo},
  journal={IEEE Robotics and Automation Letters}, 
  title={START: Traversing Sparse Footholds With Terrain Reconstruction}, 
  year={2026},
  volume={11},
  number={2},
  pages={2194-2201},
  doi={10.1109/LRA.2025.3645649}}

@article{he2025attention,
  title={Attention-based map encoding for learning generalized legged locomotion},
  author={He, Junzhe and Zhang, Chong and Jenelten, Fabian and Grandia, Ruben and B{\"a}cher, Moritz and Hutter, Marco},
  journal={Science Robotics},
  volume={10},
  number={105},
  pages={eadv3604},
  year={2025},
  publisher={American Association for the Advancement of Science}
}

@ARTICLE{dong2025marg,
  author={Dong, Yinzhao and Ma, Ji and Zhao, Liu and Li, Wanyue and Lu, Peng},
  journal={IEEE Transactions on Robotics}, 
  title={MARG: MAstering Risky Gap Terrains for Legged Robots With Elevation Mapping}, 
  year={2025},
  volume={41},
  number={},
  pages={6123-6139},
  doi={10.1109/TRO.2025.3619041}}

@article{schwarke2025rsl,
  title={Rsl-rl: A learning library for robotics research},
  author={Schwarke, Clemens and Mittal, Mayank and Rudin, Nikita and Hoeller, David and Hutter, Marco},
  journal={arXiv preprint arXiv:2509.10771},
  year={2025}
}

@article{schulman2017proximal,
  title={Proximal policy optimization algorithms},
  author={Schulman, John and Wolski, Filip and Dhariwal, Prafulla and Radford, Alec and Klimov, Oleg},
  journal={arXiv preprint arXiv:1707.06347},
  year={2017}
}

@article{hochreiter1997long,
  title={Long short-term memory},
  author={Hochreiter, Sepp and Schmidhuber, J{\"u}rgen},
  journal={Neural computation},
  volume={9},
  number={8},
  pages={1735--1780},
  year={1997},
  publisher={MIT press}
}

@inproceedings{miki2022elevation,
  title={Elevation mapping for locomotion and navigation using gpu},
  author={Miki, Takahiro and Wellhausen, Lorenz and Grandia, Ruben and Jenelten, Fabian and Homberger, Timon and Hutter, Marco},
  booktitle={2022 IEEE/RSJ International Conference on Intelligent Robots and Systems (IROS)},
  pages={2273--2280},
  year={2022},
  organization={IEEE}
}

@inproceedings{zhang2024learning,
  title={Learning agile locomotion on risky terrains},
  author={Zhang, Chong and Rudin, Nikita and Hoeller, David and Hutter, Marco},
  booktitle={2024 IEEE/RSJ International Conference on Intelligent Robots and Systems (IROS)},
  pages={11864--11871},
  year={2024},
  organization={IEEE}
}

@article{rudin2025parkour,
  title={Parkour in the wild: Learning a general and extensible agile locomotion policy using multi-expert distillation and RL fine-tuning},
  author={Rudin, Nikita and He, Junzhe and Aurand, Joshua and Hutter, Marco},
  journal={arXiv preprint arXiv:2505.11164},
  year={2025}
}

@article{ma2025learning,
  title={Learning coordinated badminton skills for legged manipulators},
  author={Ma, Yuntao and Cramariuc, Andrei and Farshidian, Farbod and Hutter, Marco},
  journal={Science robotics},
  volume={10},
  number={102},
  pages={eadu3922},
  year={2025},
  publisher={American Association for the Advancement of Science}
}

@inproceedings{miki2024learning,
  title={Learning to walk in confined spaces using 3d representation},
  author={Miki, Takahiro and Lee, Joonho and Wellhausen, Lorenz and Hutter, Marco},
  booktitle={2024 IEEE International Conference on Robotics and Automation (ICRA)},
  pages={8649--8656},
  year={2024},
  organization={IEEE}
}

@article{bloesch2017iterated,
  title={Iterated extended Kalman filter based visual-inertial odometry using direct photometric feedback},
  author={Bloesch, Michael and Burri, Michael and Omari, Sammy and Hutter, Marco and Siegwart, Roland},
  journal={The International Journal of Robotics Research},
  volume={36},
  number={10},
  pages={1053--1072},
  year={2017},
  publisher={SAGE Publications Sage UK: London, England}
}

@article{bloesch2013state,
  title={State estimation for legged robots-consistent fusion of leg kinematics and IMU},
  author={Bloesch, Michael and Hutter, Marco and Hoepflinger, Mark A and Leutenegger, Stefan and Gehring, Christian and Remy, C David and Siegwart, Roland},
  journal={Robotics},
  volume={17},
  pages={17--24},
  year={2013},
  publisher={The MIT Press}
}

@inproceedings{zhong2023touching,
  title={Touching a nerf: Leveraging neural radiance fields for tactile sensory data generation},
  author={Zhong, Shaohong and Albini, Alessandro and Jones, Oiwi Parker and Maiolino, Perla and Posner, Ingmar},
  booktitle={Conference on Robot Learning},
  pages={1618--1628},
  year={2023},
  organization={PMLR}
}

@inproceedings{patel2020deep,
  title={Deep tactile experience: Estimating tactile sensor output from depth sensor data},
  author={Patel, Karankumar and Iba, Soshi and Jamali, Nawid},
  booktitle={2020 IEEE/RSJ International Conference on Intelligent Robots and Systems (IROS)},
  pages={9846--9853},
  year={2020},
  organization={IEEE}
}

@article{makoviychuk2021isaac,
  title={Isaac gym: High performance gpu-based physics simulation for robot learning},
  author={Makoviychuk, Viktor and Wawrzyniak, Lukasz and Guo, Yunrong and Lu, Michelle and Storey, Kier and Macklin, Miles and Hoeller, David and Rudin, Nikita and Allshire, Arthur and Handa, Ankur and others},
  journal={arXiv preprint arXiv:2108.10470},
  year={2021}
}

@article{mittal2025isaac,
  title={Isaac lab: A gpu-accelerated simulation framework for multi-modal robot learning},
  author={Mittal, Mayank and Roth, Pascal and Tigue, James and Richard, Antoine and Zhang, Octi and Du, Peter and Serrano-Munoz, Antonio and Yao, Xinjie and Zurbr{\"u}gg, Ren{\'e} and Rudin, Nikita and others},
  journal={arXiv preprint arXiv:2511.04831},
  year={2025}
}
